# CRED: A Deep Residual Network of Convolutional and Recurrent Units for Earthquake Signal Detection

S. Mostafa Mousavi, Weiqiang Zhu, Yixiao Sheng, Gregory C. Beroza

Department of Geophysics, Stanford University, USA
mmousavi@stanford.edu

**Abstract**

Earthquake signal detection is at the core of observational seismology. A good detection algorithm should be sensitive to small and weak events with a variety of waveform shapes, robust to background noise and non-earthquake signals, and efficient for processing large data volumes. Here, we introduce the Cnn-Rnn Earthquake Detector (CRED), a detector based on deep neural networks. The network uses a combination of convolutional layers and bi-directional long-short-term memory units in a residual structure. It learns the time-frequency characteristics of the dominant phases in an earthquake signal from three component data recorded on a single station. We train the network using 500,000 seismograms (250k associated with tectonic earthquakes and 250k identified as noise) recorded in Northern California and tested it with a F-score of 99.95. The robustness of the trained model with respect to the noise level and non-earthquake signals is shown by applying it to a set of semi-synthetic signals. The model is applied to one month of continuous data recorded at Central Arkansas to demonstrate its efficiency, generalization, and sensitivity. Our model is able to detect more than 700 microearthquakes as small as -1.3 ML induced during hydraulic fracturing far away than the training region. The performance of the model is compared with STA/LTA, template matching, and FAST algorithms. Our results indicate an efficient and reliable performance of CRED. This framework holds great promise in lowering the detection threshold while minimizing false positive detection rates.

## 1. Introduction

During the past 10 years, there has been an enormous increase in the volume of data being generated by the seismological community. Each year, more than 50 terabytes of seismic data are archived at the Incorporated Research Institutions for Seismology (IRIS) alone. The massive amount of data highlights the need for more efficient and powerful tools for data processing and analyses. The main challenge is the efficient extraction of as much useful information as possible from these large datasets. This is where rapidly evolving machine learning (ML) approaches have the potential to play a key role (Zhu and Beroza 2018; Li et al, 2018; Ross et al, 2018b; Chen 2018 ).

One of the first stages that observational seismologists need to meet this challenge is in the processing of continuous data to detect earthquake signals. Among a large variety of detection methods developed in past few decades, STA/LTA (Allen, 1978) and template matching (Gibbons and Ringdal 2006; Shelly et al. 2007; Ross et al, 2017; Li et al, 2018) are the most commonly used algorithms. While STA/LTA is generalized and efficient, its sensitivity to time-varying background noise and lack of sensitivity to small events, false positives, and events recorded shortly after each other make it less than optimal for robust and sensitive detection. Although the high sensitivity of cross-correlation improves the detection threshold of template matching, the requirement of prior knowledge of templates and multiple cross-correlation procedures make it less general and inefficient for real-time processing of large seismic data volumes. Although more advanced algorithms such as Fingerprint And Similarity Thresholding (FAST) (Yoon et al, 2015) can improve the efficiency of the similarity search, the outputs are in that case limited to repeated events.

Shallow Neural Networks (NN) are among the first ML methods used for the earthquake signal detection (e.g. Zhao and Takano 1999; Wang and Teng, 1997; Madureira, and Ruano, 2009). NN receive a feature vector, x (a sparse representation of seismic data), as input and transform it through a series of hidden layers to predict the desired outputs, y, in the output layer. Each hidden layer is made up of a set of neurons, where each neuron is fully connected to all neurons in the adjacent layers, and where neurons in a single layer function completely independently and do not share any connections. Non-linear activation functions inside neurons make learning complex x-y relations possible through an optimization process. To learn these relations more deeply and to build a more complex model requires exponentially more units in each layer or alternatively many layers with few units in each layer (deep neural net). Fully connected layers, as in standard neural networks, don't scale well to large input vectors and require careful initiation, feature engineering, optimization, and regularization to prevent vanishing/exploding gradients and overfitting problems.

During the recent renewed interest in ML applications among seismologists, some studies have revisited the detection problem (e.g. Perol et al., 2018; Wu et al., 2018; Ross et al. 2018a). These studies presented successful and promising performances of deep convolutional-neural-networks (CNN) for robust and efficient detection of earthquake signals. In addition to fully connected layers, CNN's include convolutional layers and pooling layers. A convolutional layer consists of a set of learnable filters (or kernels). Every filter has a small receptive field (connected to a small region of the layer before it). It applies a convolution operation to the input and produces an activation map that gives the responses of that filter at every spatial position. Hence, as the input volume is passing through the convolutional layers the network automatically learns the low/mid/high-level features in data. This eliminates the need for heavy pre-processing and hand engineering of features. Moreover, it reduces the number of free parameters, allowing the network to be deeper with fewer parameters. Pooling layers combine the outputs of multiple neurons at one layer into a single neuron in the next layer, hence it can reduce the dimension of learned features as data is fed forward.

In this paper, we formulate the detection problem as a sequence-to-sequence learning (Sutskever et al. 2014) where a time series of inputs are mapped to a time series of probability outputs and separate predictions were made for each individual sample. Recurrent neural networks (RNN) were designed for processing sequential data like seismograms. They have internal state (memory) and can share features learned across the different positions within a time series (Bayer et al., 2009). Hence, they can learn the dynamic temporal relations within a time sequence. In this study, we use bi-directional long-short-term memory (LSTM) as our RNN unites and employ a residual-learning framework to make a deeper learning feasible. In the residual learning framework, the network will learn the residual functions instead of original mapping functions. This makes their optimization easier, keeps the training error in the deeper layer the same as for the shallower ones, and by doing so makes it possible to train a deeper network. Deeper networks allow learning more high-level features and building more complex models. We demonstrate the performance of the network using both semi-synthetic and real data.

## 2. Methodology

2.1. Sequential Learning
The RNN performs sequential learning by retaining relations among inputs while training itself (Figure 1).

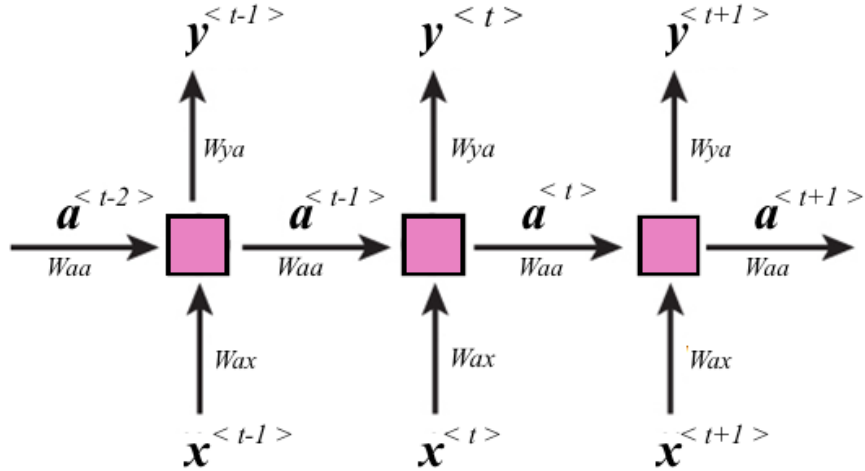

**Figure 1.** Schematic of RNN for many-to-many elements. Internal states, *a*, are passing through the network and at each position used for calculating the prediction *y*.

A nonlinear function of the weighted sum of input at time *t*, $x^{<t>}$, and the previous state (or learned activation in the previous time step), $a^{<t-1>}$, is used to compute the current state, $a^{<t>}$, and predict the output at time *t*, $y^{<t>}$.

$$a^{<t>} = gt(W_{aa}a^{<t-1>} + W_{ax} x^{<t>} + b_a) \qquad (1)$$

$$y^{<t>} = gs(W_{ya}a^{<t>} + b_y) \qquad (2)$$

where the *W*'s and *b*'s are associated weights and bias terms, and *gt* and *gs* are Tanh and Sigmoid activation functions:

$$gt = \frac{e^z - e^{-z}}{e^z + e^{-z}} \qquad (3)$$

$$gs = \frac{1}{1 + e^{-z}} \qquad (4)$$

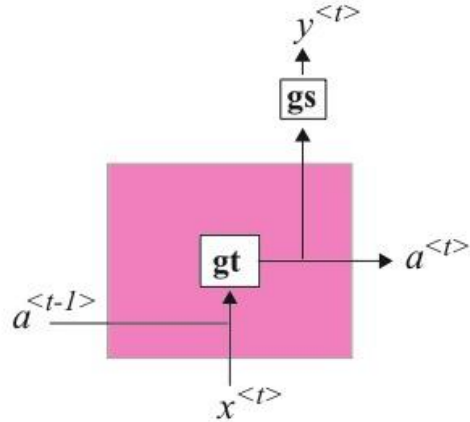

**Figure 2.** A closer look at a standard RNN unit in the previous figure. Inside each unit, a Tanh activation function, *gt,* is applied to the internal states from the previous position and the input of current position to obtain the internal state, *a,* at current position. This state then will be passed to the next position and also will be used to predict the output of the current position, *y,* using either a Sigmoid or SoftMax function, *gs*.

This basic RNN unit (Figure 2), however, is not effective for learning long sequences due to the vanishing/exploding gradient problem. Some specific types of RNN such as long-short-term memory (LSTM) are commonly used to reduce the vanishing/exploding gradient problem and make application of deeper networks feasible.

An LSTM (Hochreiter and Schnidhuber 1997) unit has an internal memory cell, c, which is simply added to the processed input. This greatly reduces the multiplicative effect of small gradients. The time dependence and effects of previous inputs are controlled by a forget gate, which determines what states are remembered or forgotten. Two other gates, the update gate, and an output gate are also featured in LSTM cells (Figure 3).

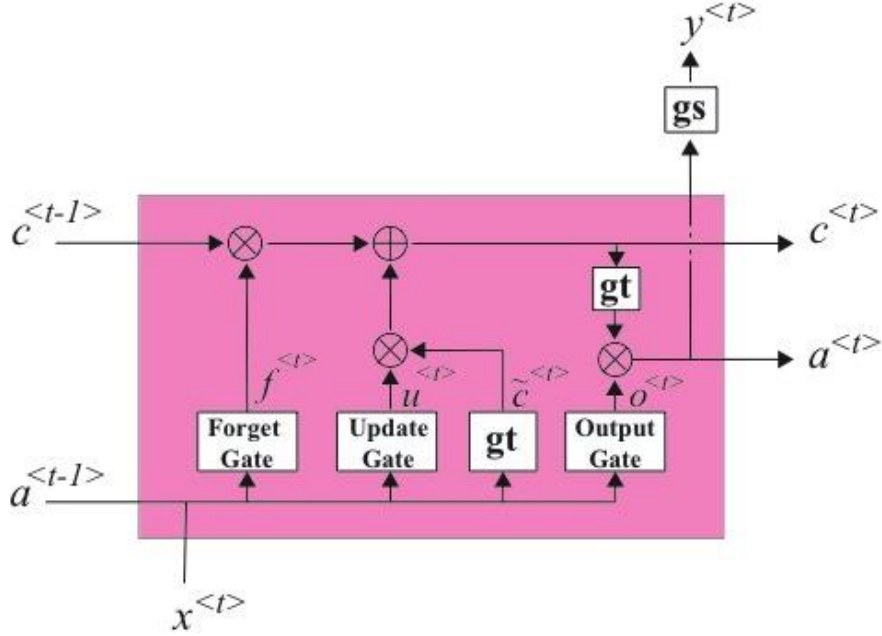

**Figure 3.** An LSTM unit. These units in addition to the long-term memory, *a,* have a short-term memory c.

$$c^{\sim <t>} = gt(W_c a^{<t-1>} + W_c x^{<t>} + b_c) \tag{5}$$

$$u^{<t>} = gs(W_u a^{<t-1>} + W_u x^{<t>} + b_u) \tag{6}$$

$$f^{<t>} = gs(W_f a^{<t-1>} + W_f x^{<t>} + b_f) \tag{7}$$

$$o^{<t>} = gs(W_o a^{<t-1>} + W_o x^{<t>} + b_o) \tag{8}$$

where $c^{\sim <t>}$ is candidate value for replacing the memory, $u^{<t>}$ is update gate, $f^{<t>}$ is forget gate and $o^{<t>}$ is the output gate. The value of memory cell at each time step will be set using the candidate value at current step ($c^{\sim <t>}$) and previous value ($c^{<t-1>}$) based on update and forget gates:

$$c^{<t>} = u^{<t>} * c^{\sim <t>} + f^{<t>} * c^{<t-1>} \tag{9}$$

where * is element-wise multiplication. The final state at time t, $a^{<t>}$, is obtained based on output gate and the value of memory cell:

$$a^{<t>} = o^{<t>} * gt(c^{<t>}) \tag{10}$$

These LSTM units can be connected to each other in a row to form one hidden layer of the RNN (Figure 4). To learn the temporal pattern both from left to right and right to left, we can simply concatenate two of these layers with opposite directionality to form a bidirectional (Thireou and Recsko 2007) RNN layer (Figure 4). This structure allows the networks to have both backward and forward information about the sequence at every time step.

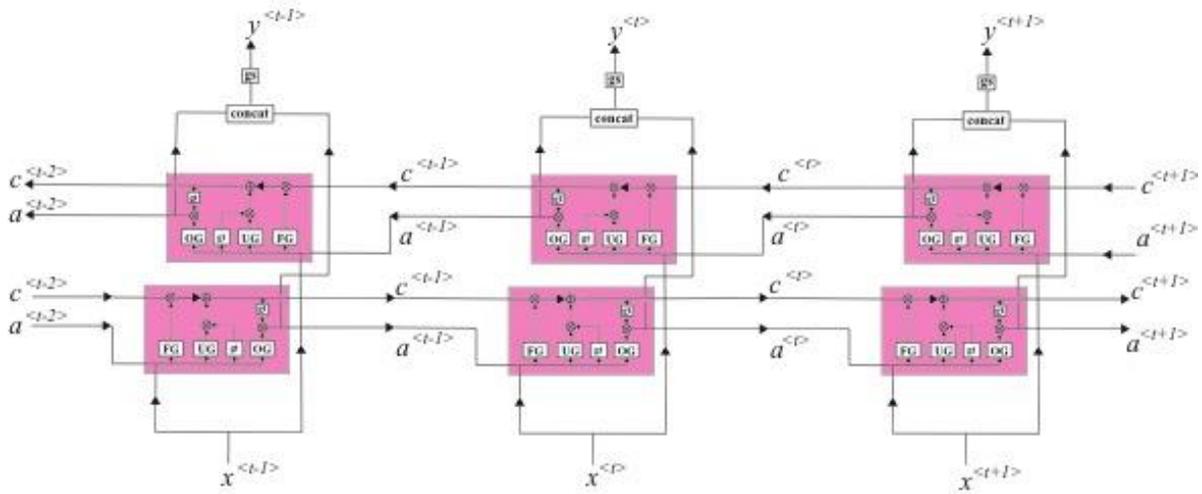

**Figure 4:** bidirectional LSTM layer

2.2. Residual Learning

In an end-to-end deep network, the "levels" of learned features can be enriched by the number of stacked layers (depth). For that reason, the number of layers (depth of a network) plays a crucial role in allowing the network to learn more high-level features and build more complex models. There are, however, two main problems that occur in the training of very deep networks (where "deep" is usually taken to mean more than 15 layers): (1) vanishing/exploding gradients and (2) degradation (saturation and degradation of accuracy). The first problem can be largely addressed by normalized initialization. An effective solution for the later was proposed by He et al (2016) by introducing the residual learning framework.

In a standard deep network, every few stacked layers basically learn a nonlinear underlying function, H, that maps the input to the first of these layers, x, to the output of the final layer, H(x). In the residual learning framework, the original mapping is recast into F(x) + x. Each few stacked layers (a residual block) rather than directly learning H(x), learns the residual functions, F(x) := H(x) − x. Since optimization of the residual mapping is easier than for the original unreferenced mapping, this can help to keep the training error in the deeper layer the same as for the shallower ones, and by doing so make it possible to train a deeper network. Residual learning blocks can be realized by feedforward neural networks with "shortcut connections" (Fig. 5). Shortcut connections are those that skip one or more layers. The shortcut connections simply perform identity mapping, and their outputs are added to the outputs of the stacked layers (Fig. 5). These identity shortcut connections add neither extra parameters nor computational complexity to the problem.

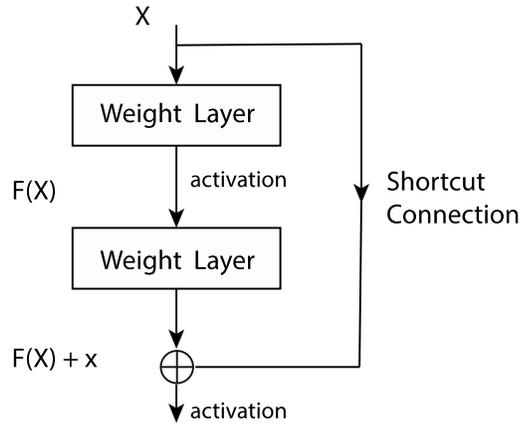

**Figure 5.** A building block in Residual learning.

In residual learning a building block is defined as:

$$y = F(x, \{W_i\}) + x. \qquad (11)$$

Here x and y are the input and output vectors of the layers considered. The function $F(x, \{W_i\})$ represents the residual mapping to be learned. For the example in Figure 5 that has two layers, $F = W_2\sigma(W_1 x)$ in which σ denotes the activation function and the biases are omitted for simplifying notations. The operation $F + x$ is performed by a shortcut connection and element-wise addition. As noted above, the shortcut connections introduce neither extra parameters nor computation complexity.

2.3. The Network Architecture

The architecture of our proposed network is presented in Figure 6. Three types of convolutional, recurrent, and fully connected layers have been used in a residual structure.

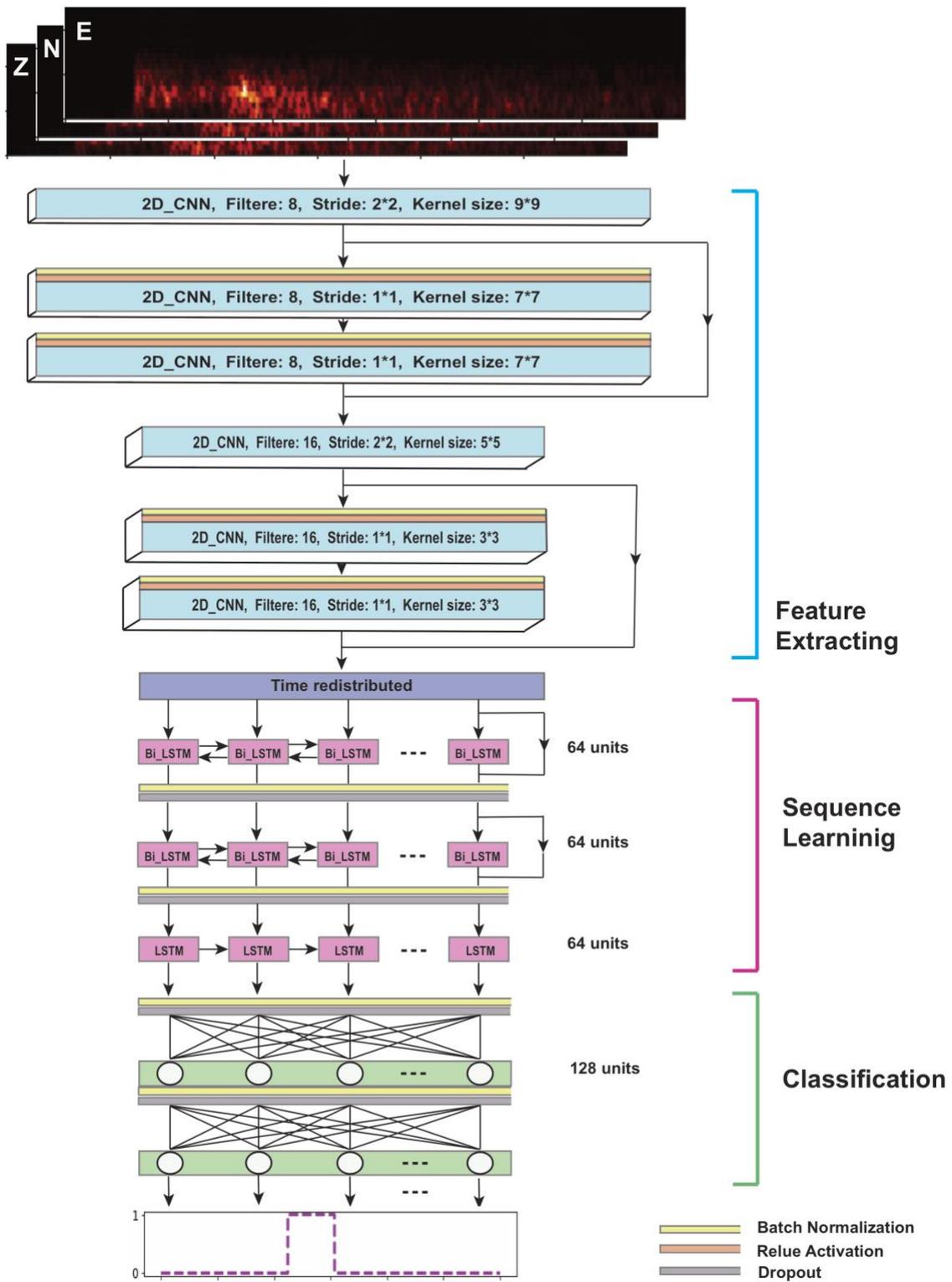

**Figure 6.** The architecture of our proposed very-deep neural network.

Inputs into the network are spectrograms of three component seismograms. Our previous studies (Mousavi et al. 2016a,b; Mousavi and Langston 2016a,b) showed that time-frequency transforms can improve event detection by better characterizing seismic signals. The first few blocks (blue layers) consist of mainly convolutional layers. Convolutional layers in the residual block are preceded by a batch normalization (BN) and an activation layer. BN layers normalize the activation of a convolution layer during the training, which accelerates the training and has some regularization characteristics that help prevent overfitting. These layers will automatically learn and extract features from the input spectrograms so there is no need for feature engineering or complicated pre-processing in advance. As the network gets deeper, these layers extract more high-level (i.e., finer scale) features.

To maximize the feature extraction we increased the number of 2D convolutional layers by arranging them into residual blocks based on the improved structure presented in He et al (2017). This is done by providing shortcuts that make it possible to train a deeper network without degradation. Every two residual blocks are preceded by an individual layer with a stride of 2 that halves the feature map size. This will decrease the feature map size with depth and down-sample the feature volume to a more sparse representation. The number of filters in every three layers will be doubled to preserve the time complexity per layer. The down-sampling will speed up the training of recurrent layers in the following section.

After 2D convolutional layers, the feature volume will be redistributed into a sequence and will pass into two residual blocks of bidirectional LSTMs. These layers will learn and model the sequential pattern of data. Since in real time seismic acquisition samples of an earthquake signal are recorded/represented with time increasing from left-to-right, we add one additional unidirectional LSTM layer to the end of the recurrent section. Finally, the last two layers of the network are fully connected layers that perform the high-level reasoning and map the learned sequence model from the last step to the desired output classes. A sigmoid binary activation function was used in the last layer of the network and the output is a vector of predicted probabilities for each sample to contain an earthquake signal.

Overall the network consists of 12 layers and has 256,000 trainable parameters. It takes advantage of three different types of layers within an efficient structure. Design of the network with shortcuts in a residual learning structure makes it possible to train the network with an acceptably low error rate. This is an end-to-end learning framework that can characterize the seismic data with high precision. It is designed to be extensible and has the potential to fulfill many applications in automating seismic data processing.

### 3. Data

550,000 30-second 3-component seismograms recorded by 889 broadband and short-period stations in North California are used for the training of the network and its validation. 50% of these seismograms are associated with earthquakes occurring from January 1987 to December 2017. The metadata used for the labeling *P*-wave and *S*-wave arrival times are manual picks provided by the Northern California Earthquake Data Center. Another half of the data set consists of seismic noise recorded by the same network stations. The noise samples contain a variety of ambient and non-ambient noises (Figure 7). 30-second seismograms were randomly obtained for the time spans between the cataloged events. Then a simple detection algorithm was used to eliminate those traces containing uncataloged events. To further ensure no contamination of the noise samples by small events behind the background noise level, we used a de-signaling algorithm (Mousavi and Langston 2017) on the remaining traces. This algorithm removes the anomalous spectral features associated with earthquake signals.

All the traces were then detrended by removing the mean, band-pass filtered between 1 and 45 Hz, resampled at 100 HZ, and normalized. The Short Time Fourier Transform was used to construct spectrograms. For ground truth, we generated one binary label vector with the same length as spectrograms. In the label vector, we set the value of corresponding samples form P arrival to P + 3d (d= S-P) to 1 and the rest to 0 representing the probabilities (Figure 7). This range can capture the dominant spectral energy of P and shear waves which characterize well the earthquake signals in the time-frequency domain.

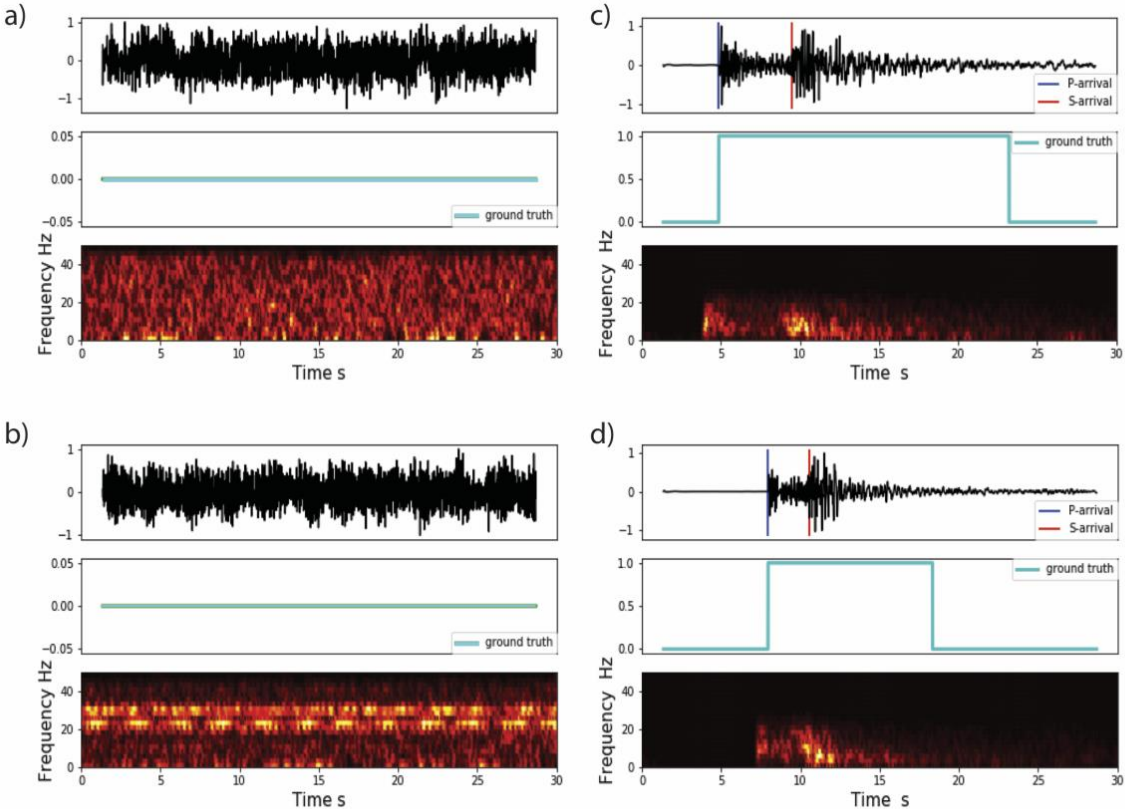

**Figure 7.** Examples of seismogram, label vector, and associated spectrogram (short time Fourier transform, STFT) for vertical components of two sample noises (a and b) and two earthquake samples (c and d).

### 4. Training and Testing of the Model

We randomly split the dataset into training (80 %), evaluation (10 %), and test (10 %) sets. We trained the network on one Tesla V100-PCIE GPU. Binary cross-entropy was used as the loss function and the ADAM algorithm (Kingma and Ba, 2014) was used for optimization. The training was completed in 62 epochs, and we found that the validation accuracy did not improve in the last 20 epochs. The final training and validation accuracies are 99.33 % and 99.24 % respectively while the losses are 0.02 and 0.03. The mean absolute errors are 0.01 and 0.009 for the training set and validation set respectively. These values are measured from a point-by-point comparison of the model predictions with the ground truth.

To evaluate the performance of the model for detection, we used 50,000 test samples. After selecting a decision threshold value (tr) for output probabilities, several parameters for evaluating the performance of the model can be calculated. We use precision, recall, and F-score as evaluation metrics. Precision is defined as the fraction of predictions that are accurate, Recall is defined as the fraction of instances that are accurately predicted, and F-score combines these two parameters to eliminate effects of unbalanced sample size for different classes:

$$\text{Precision} = \frac{\text{TP(tr)}}{\text{TP(tr)}+\text{FP(tr)}} \tag{12}$$

$$\text{Recall} = \frac{\text{TP(t)}}{\text{TP(tr)}+\text{FN(tr)}} \tag{13}$$

$$\text{F} - \text{score} = \frac{2\times\text{Precision}\times\text{Recall}}{\text{Precision}+\text{Recall}} \tag{14}$$

where, TP denotes true positives, FP denotes false positives, and FN are false negatives. In a perfect classifier *TP=1,* and *FP=0*. The precision and recall are calculated using different threshold values to check the effect of different threshold selection on the detection results (Figure 8). As can be seen from Figure 8, regardless of the threshold choice, the model test results in a precision higher than 96 % and a recall above 99 %. Threshold values between 0.1 to 0.3, however, lead to the maximum F-score of 99.95. In Table 1 we report the confusion matrix for the threshold value of 0.11.

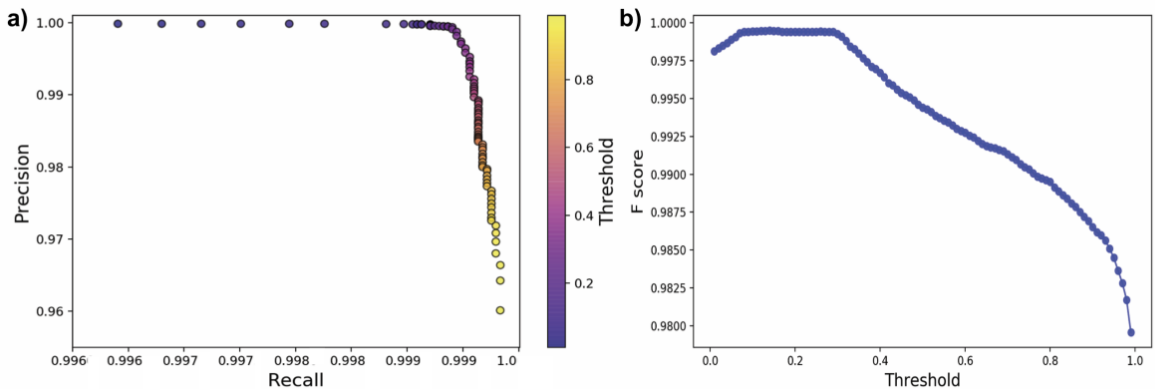

**Figure 8.** precision-recall curve (a) and the F-score as a function of threshold values (b).

**Table 1.** Confusion Matrix for Threshold = 0.11

| Detected as-> | Earthquake | Noise |
|---|---|---|
| Earthquake | 25,226 | 6 |
| Noise | 23 | 24,745 |

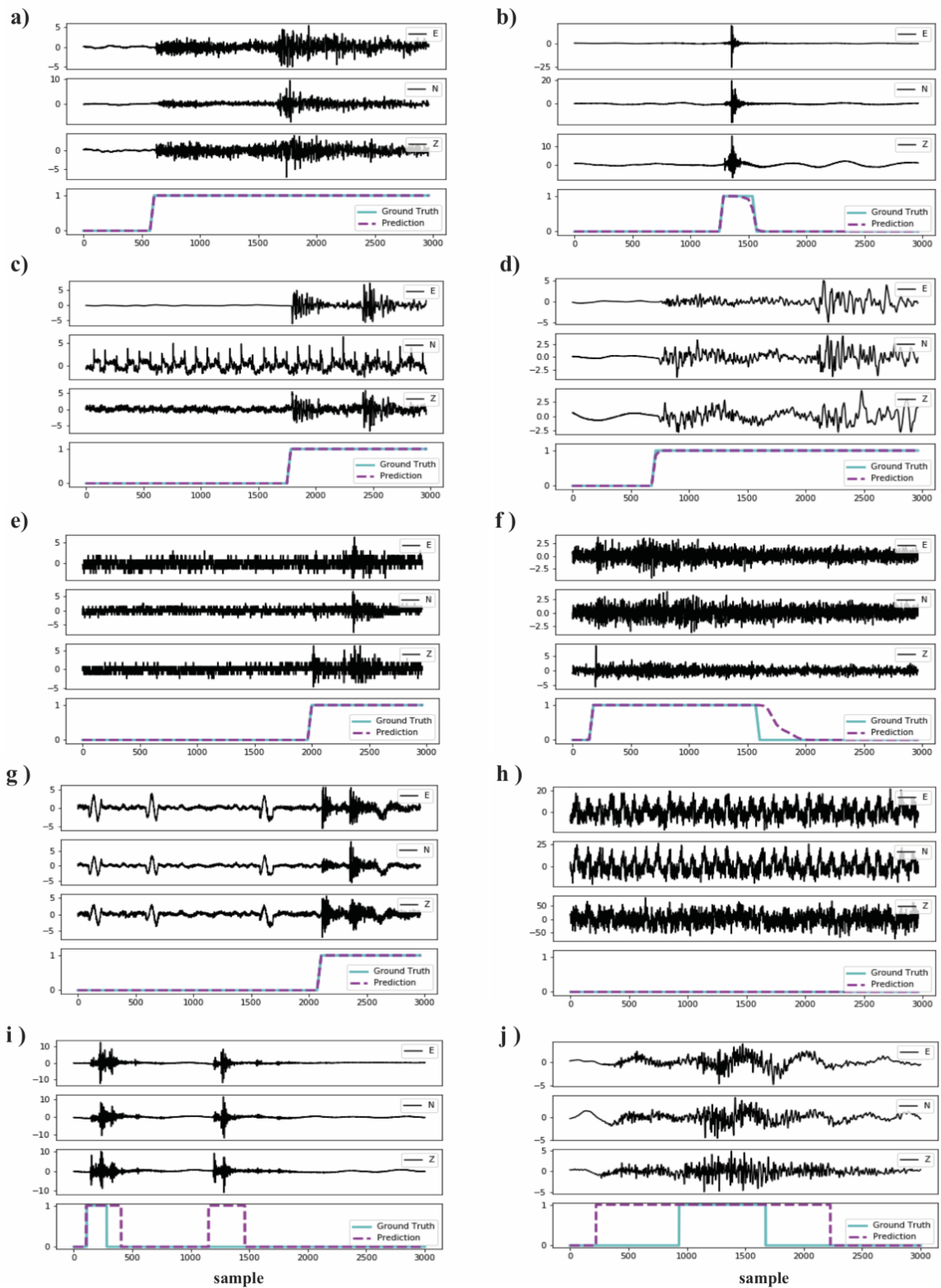

**Figure 9.** Results of applying the trained network on a few samples in the test set.

From Figure 9 we see that the network was able to generalize its learning and predict the entire duration of earthquake signals (including S coda) with high precision no matter if the event is large (Figure 9a) or small (Figure 9b), local (Figure 9b) or teleseismic (Figure 9d), recorded on an instrument with broken channel (Figure 9c), old stations (Figure 9e), contaminated with high background noise level (Figure 9f), or not earthquake signals (Figure 9g). Since the predictions are done for each sample individually, the network does not require the full length of a signal to detect an earthquake. This is important for real-time processing. In Figure 9i we see that it was able to detect an event that was missing from the catalog we used for the labeling, or correct the mislabeled P arrival and estimated end of coda in Figure 9j, and by doing so, in these instances it improved upon "ground truth."

### 5. Sensitivity Test

To explore the performance sensitivity of the CRED to the background noise level and compare it with the other popular algorithms, we performed a test using semi-synthetic data. We selected 500 events with high-SNR from the test set. Each seismogram was visually checked prior to the test. We also generated 500 Ricker wavelets with random width and scaled amplitudes as non-earthquake signals. Earthquake and non-earthquake signals were then randomly assembled to generate 8.4 hours of continuous data. Next, 23 different levels of Gaussian noise were added to the continuous waveforms in a way that result in 23 realizations of data with SNRs ranging from -2 dB to 20 dB. SNR is measured as $10*\log_{10}([S_A/N_A]^2)$ where $S_A$ and $N_A$ are peak amplitudes of signal and noise respectively. A portion of the generated synthetic data is presented in Figure 10.

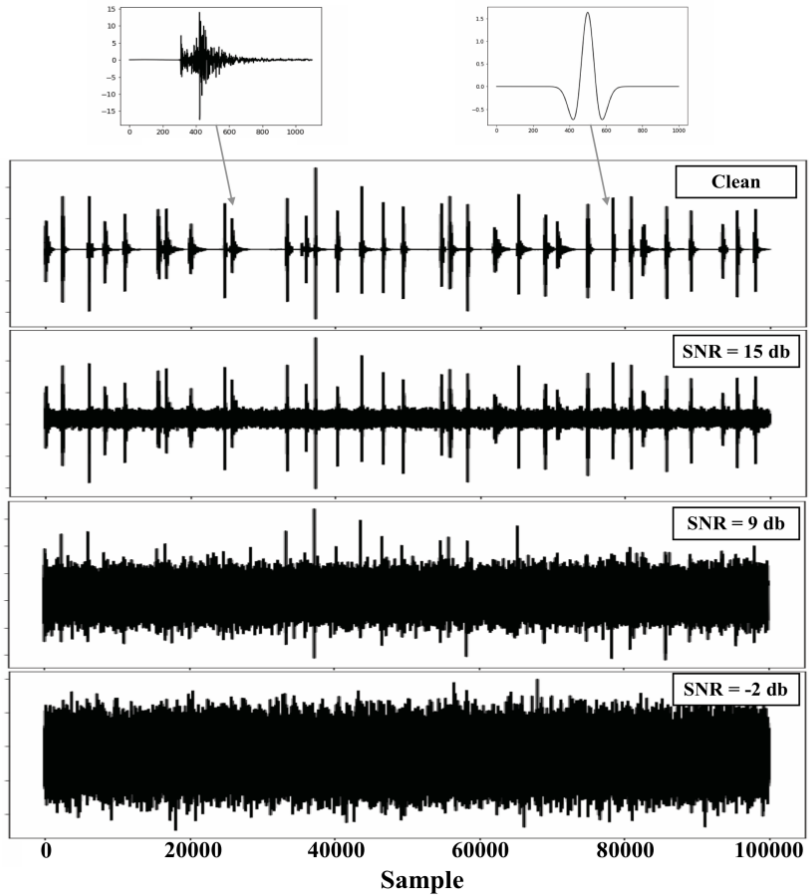

**Figure 10.** 1000 seconds of the generated synthetic seismogram (vertical component) with the different noise level. The insets at the top present one example of the earthquake and non-earthquake signals.

We then applied the CRED, STA/LTA, and template matching on the synthetic data. For the template matching two templates were used. The detection threshold for each of these algorithms was tuned carefully to maximize the precision. The results are presented in Figure 11.

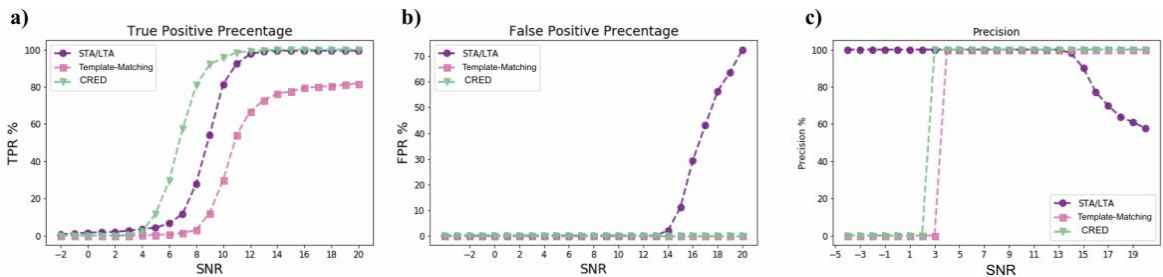

**Figure 11.** Comparison of the performance of the proposed method (CRED) with STA/LTA and template matching on semi-synthetic data with different noise levels.

As can be seen from the Figure 11, CRED had superior performance compared to two other methods. It detects 100 % events with SNR >= 12 db with 0 false positives for all SNRs. At SNR of 7 db, CRED detected 80 % of events while the detection rate for STA/LTA and template matching at this level were 27% and 3% respectively. In general CRED can detect far more events at low SNRs with 0 false positive rate. Template matching also resulted in 0 false positives but it had the lowest detection rate (81.8 % at 20 db) due to the fact that most of the events were not located nearby or recorded on the same station. On the other hand STA/LTA resulted in the highest false positive rate (72.6 % at 20 db). This is due to the sensitivity of its characteristic function to the abrupt amplitude changes, which causes a low precision for the STA/LTA at high SNRs (Figure 11c). The superior performance of CRED in noisy condition can be due to its reliance on spectral contents of the signal rather than the waveform.

## 6. Application to Central U.S.

A good model that does not overfit the data set used for the training, should generalize well to other data sets. To test this characteristic of our detector and the obtained model using the North California data set, we apply CRED to continuous data recorded in Guy-Greenbrier, Arkansas, during the 2011 sequence (Horton 2012, Mousavi et al. 2017). Several earthquake catalogs exist for this sequence based on different detection methods: Horton (2012) (ANSS) and Ogwari et al. (2016) based on STA/LTA, Huang and Beroza, (2015) based on templet matching, and Yoon et al., (2017) based on FAST. August 2010 is the overlapping time period in these catalogs, hence we made a unified catalog for this period combining information of all the above catalogs containing 3788 events. Most of these events are microearthquakes with local magnitudes ranging from -1.3 to 0.5 associated with hydraulic fracturing or wastewater injection. Events are located within a 2 km area at 2 to 4 km depths. Template matching (Huang and Beroza, 2015) has the highest detection rate (3732), FAST (Yoon et al., 2017) detected 3266 of the events, and studies based on STA/LTA (Horton 2012 and Ogwari et al. 2016) detected 23 and 24 events respectively.

We process one month of continuous data recorded at station WHAR during August 2010. This is the common station used in these studies. Detecting induced microearthquakes in different regions and at the local scale in the presence of high noise levels is a challenging task for a detection algorithm. We applied the CRED algorithm using a moving window of 30 seconds. The total processing time including transferring three-channel data into STFT and applying the model was 1 hour and 9 minutes on a laptop with a 2.7 GHz Intel Core i7 processor and 16 GB of memory. Our model detected 1102 events. Of these, 680 events were already in the catalog. We visually checked the remaining detections and verified 77 of them as new events and 345 of them as false positives. This leads to a detection precision of ~ 69 %. Detected events by CRED range from magnitude -1.2 to 2.6 (Figure 12). Lowering the detection threshold can lead to detection of more cataloged events; however, it would also result in more false positives.

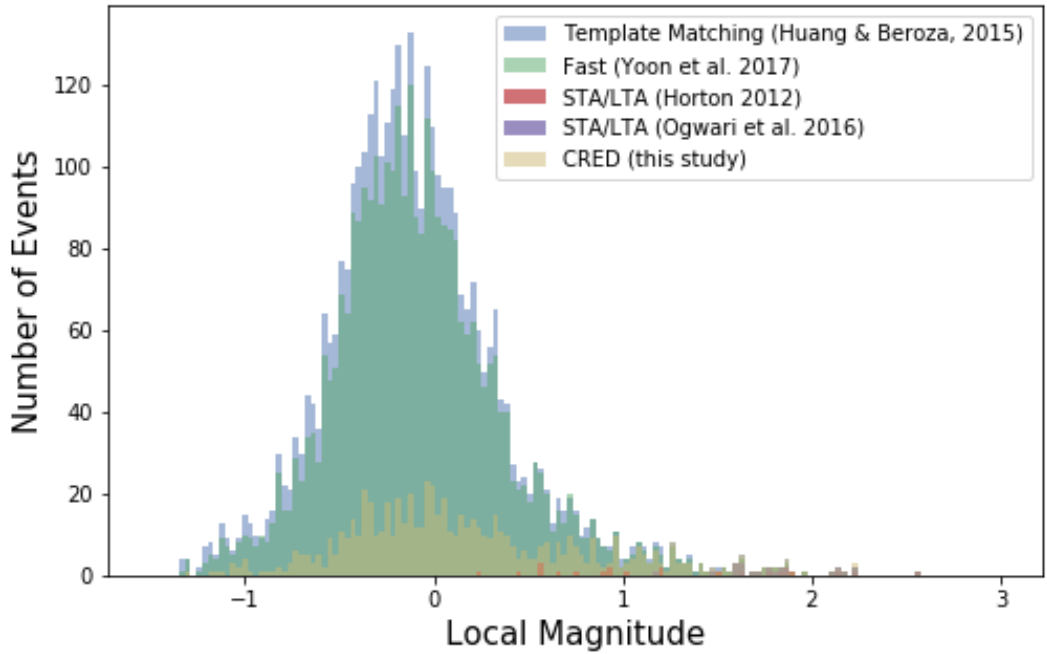

**Figure 12.** Magnitude-frequency distribution of detected events using different algorithms.

A few examples of events detected by CRED are presented in Figure 13. Almost all of the events larger than $M_L$ 1.0 have been detected. The detection extends to lower magnitudes and includes a variety of SNRs and waveforms. Magnitude calculation of newly detected events is beyond the scope of this paper; however, by comparing the amplitudes of these events with the waveforms of some of the cataloged events we estimate $0.1 < M_L < 0.4$ for these events.

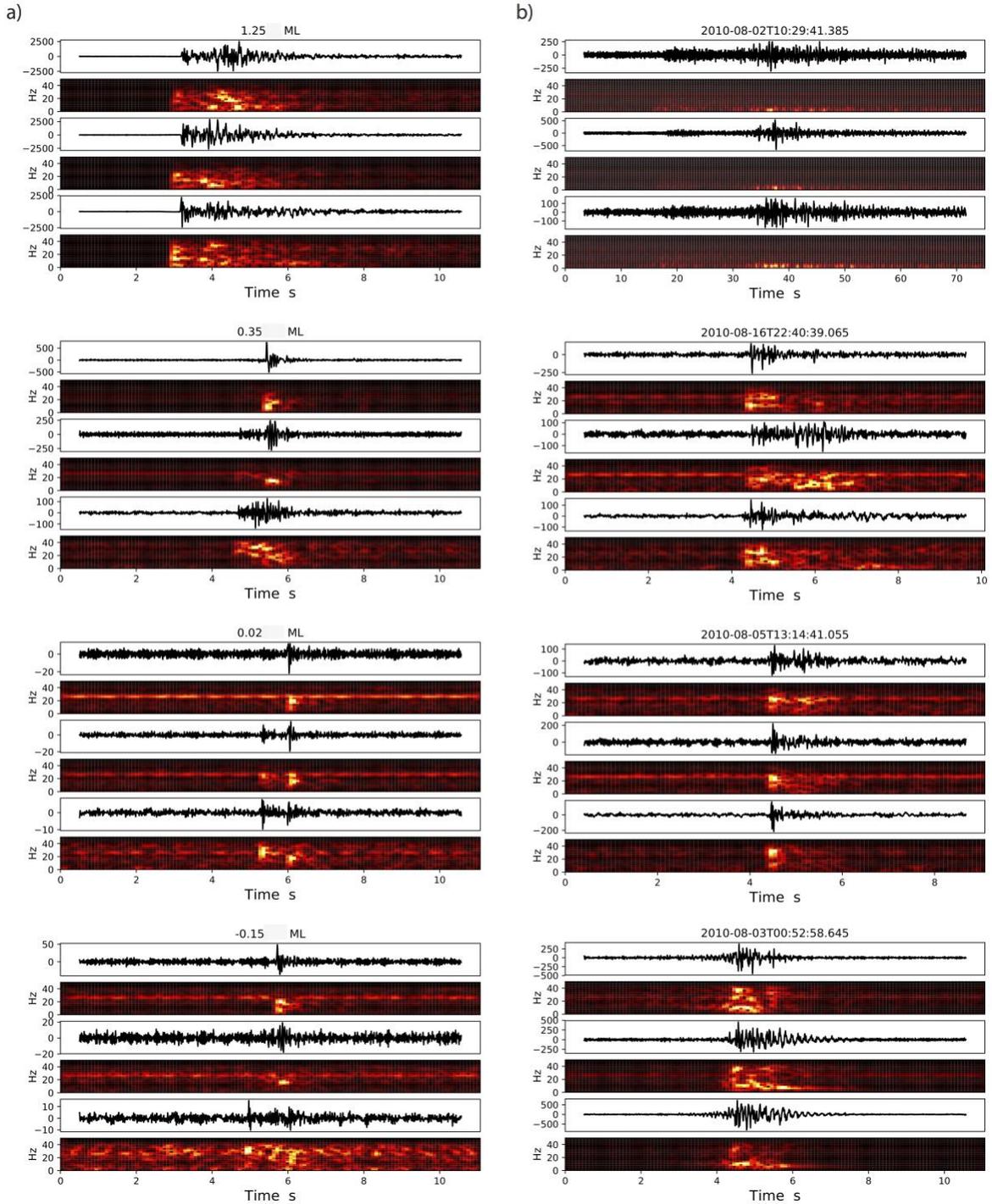

**Figure 13.** Some examples of detected events in Arkansas. Left column (a) are events that have been detected by previous studies and existed in our unified catalog. The magnitude of each event is taken from Yoon et al. (2017) and is listed on top of each plot. The right column (b) shows some events newly detected by CRED. The detection time of each event is shown on the plot headers.

## 7. Discussion

Reliable earthquake signal detection is at the core of observational earthquake seismology. While improving the sensitivity and robustness of current algorithms is still an active field of research, improving the efficiency has become the center of attention in the recent years due to a significant increase in data volumes. In this paper we explored the application of a deep residual network of convolutional and recurrent units for earthquake signal detection. Convolutional layers are powerful tools for automatic extraction of sparse features from seismograms. Recurrent units can learn sequential characteristics of seismic data and provide robust models. Here we used LSTM, a powerful type of recurrent unit to search for earthquake signals in the time-frequency domain.

We designed the network with a residual structure to prevent degradation and reach a higher accuracy with a deeper learning. The residual-learning structure of the network makes a very-deep end-to-end learning of seismic data feasible. The proposed network, can learn the dynamic time-frequency characteristics of earthquake signals and build a generalized model using a modest-sized training set. Using time-frequency representation of seismic data and the sparsity of learned features by the CNN layers make the model robust to the noise. Sensitivity tests revealed the superior performance of CRED in the presence of high noise levels compared with both STA/LTA and template matching.

The learned model generalizes well to seismic data recorded in other regions. We trained the network using local and regional waveforms recorded in Northern California. The events mainly have tectonic origins, range in 0-5 M with the highest frequency around 2 M (Figure 14), and epicentral distances of mostly around 50 km. The model was tested for detecting microseismic events in Central Arkansas, with substantially different crustal structure, much lower magnitude, shorter epicentral distance, and generally shallower depths. This represents a very challenging test of the method, but our model performed acceptably by detecting 3 orders of magnitude more events compared with STA/LTA results. Although CRED's detection rate was ~ 30 % of template matching and FAST, no templates were used in its training and it ran more than 100 times faster than FAST (non-parallel version). Detection of new events missed by all previous studies indicates the inherent limitation of similarity-search based methods even with their high computational costs.

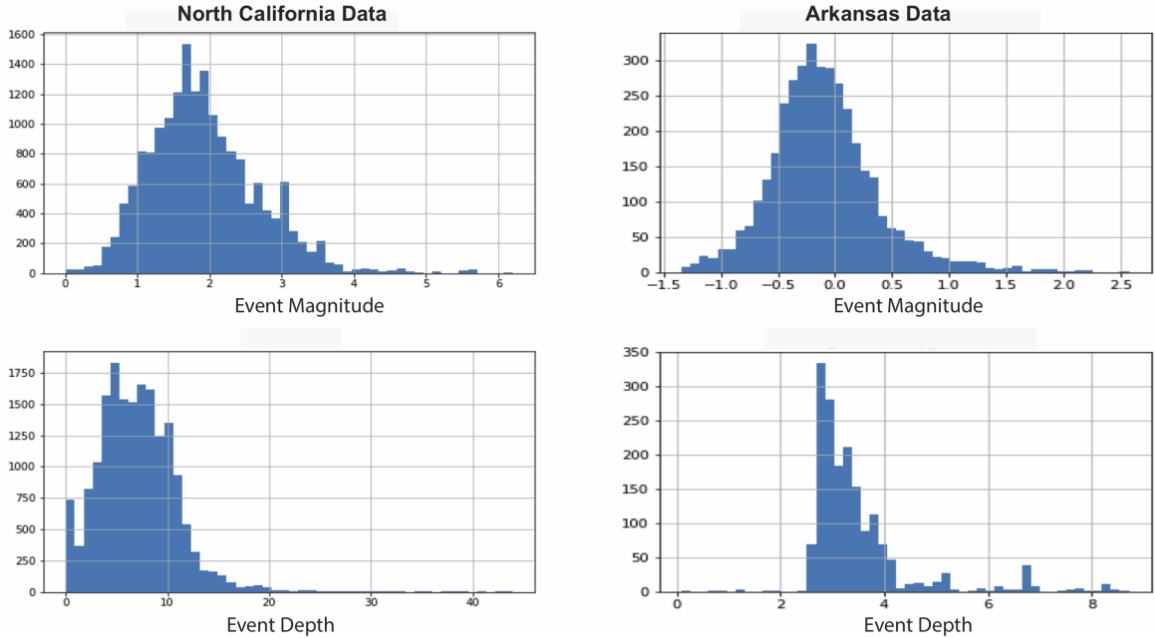

**Figure 13.** statistics of the North California Data set used for the training and the Arkansas data set.

Here we used only 250k waveforms of earthquakes with limited parameter distribution for training. Retraining the network with a much larger data set with a higher variability of earthquake signals with different hypocentral parameters and magnitude ranges should improve the performance. Moreover, a recursive approach can be adapted by the inclusion of events detected in the first round of applying the model into retraining and updating the initial model. Including more reliable labels through a secondary cross-validation step or incorporating different types of non-earthquake signals recorded on seismic stations are other steps that could be implemented to further improve the method.

A potential application of CRED is in real-time monitoring of seismic activities for hazard mitigation in tectonically active or induced seismic zones. Microseismic monitoring and earthquake early warning are other areas where CRED could be used. This framework can provide fast and reliable event detection and is easily scalable to large N (many sensor) or large T (long duration) data. Reprocessing large volumes of archived data can potentially lead to new insights into a wide variety of earthquake phenomena.

## 8. Conclusion

In this paper we present a Cnn-Rnn Earthquake Detector (CRED), which is a network that combines convolutional and recurrent units for deep residual learning of the time-frequency characteristics of earthquake signals. This framework is capable of detecting seismic events with high efficiency and precision. The learned model has low sensitivity to the background noise level, can generalized well to other regions with different seismicity characteristics, and outperforms STA/LTA in terms of sensitivity and template matching in terms of efficiency. Application of our method is fast and once the network is trained it can be applied to a stream of seismic data in real time. The architecture is flexible and can be scaled up easily. False positive rates are minimal due to the high-resolution modeling of earthquake signals based on their spectral structure.

**Acknowledgments**

We would like to thank Stephane Zuzlewski for this help in accessing the waveform data and meta data archived in NCEDC. Clara Yoon and Paul Ogwari helped us in preparing the catalogs for the Guy-Greenbrier sequence. This study was supported by the Stanford Center for Induced and Triggered Seismicity. Waveform data and metadata for the training and validation was obtained from the Northern California Data Center. The continuous data for Arkansas was downloaded from IRIS.

**References**

Bayer, Justin; Wierstra, Daan; Togelius, Julian; Schmidhuber, Jürgen (2009). "Evolving Memory Cell Structures for Sequence Learning". Artificial Neural Networks – ICANN 2009. Lecture Notes in Computer Science. Springer, Berlin, Heidelberg. 5769: 755–764. doi:10.1007/978-3-642-04277-5_76. ISBN 978-3-642-04276-8.

Chen, Y. (2018). Automatic microseismic event picking via unsupervised machine learning. Geophysical Journal International, 212(1), 88–102.

Gibbons, S. J., and Ringdal, F. (2006). The detection of low magnitude seismic events using array-based waveform correlation. Geophysical Journal International, 165(1), 149–166.

He, K., X. Zhang, S. Ren and J. Sun, (2016) "Deep Residual Learning for Image Recognition," 2016 IEEE Conference on Computer Vision and Pattern Recognition (CVPR), Las Vegas, NV, 2016, pp. 770-778, doi: 10.1109/CVPR.2016.90.

Hochreiter S. and J. Schmidhuber, (1997) "Long Short-Term Memory," Neural Computation, vol. 9, no. 8, pp. 1735-1780, Nov. 15, doi: 10.1162/neco.1997.9.8.1735.

Horton, S., 2012. Disposal of hydrofracking waste fluid by injection into subsurface aquifers triggers earthquake swarm in central Arkansas with potential for damaging earthquake. Seismol. Res. Lett. 83, 250–260.

Huang, Y., Beroza, G.C., 2015. Temporal variation in the magnitude-frequency distribution during the guy-greenbrier earthquake sequence. Geophys. Res. Lett. 42 (16), 6639–6646. http://dx.doi.org/10.1002/2015GL065170.

Kingma, D.P., and J. Ba (2014) Adam: A Method for Stochastic Optimization. arXiv: 1412.6980 [cs], Dec. 2014. arXiv: 1412.6980

Li, Z., Peng, Z., Hollis, D., Zhu, L., and McClellan, J. (2018). High-resolution seismic event detection using local similarity for Large-N arrays. Scientific Reports, 8(1), 1646.

Li, Z., Meier, M.-A., Hauksson, E., Zhan, Z., and Andrews, J. (2018). Machine Learning Seismic Wave Discrimination: Application to Earthquake Early Warning. Geophysical Research Letters, 45(10), 4773–4779.

Madureira, G., Ruano, A.E., (2009). A neural network seismic detector. Acta Technica Jaurinensis, 2 (2), 159-170.

Mousavi, S. M., S. Horton, P. Ogwari, and C. A. Langston, (2017). Spatio-temporal Evolution of Frequency-Magnitude Distribution during Initiation of Induced Seismicity at Guy-Greenbrier, Arkansas, Physics of the Earth and Planetary Interiors, 267, 53-66, http://doi.org/10.1016/j.pepi.2017.04.005

Mousavi, S. M., and C. A. Langston (2017). Automatic Noise-Removal/Signal-Removal Based on the General-Cross-Validation Thresholding in Synchrosqueezed domains, and its application on earthquake data, Geophysics.82(4), V211-V227 doi: 10.1190/geo2016-0433.1

Mousavi, S. M., C. A. Langston, and S. P. Horton (2016a). Automatic Microseismic Denoising and Onset Detection Using the Synchrosqueezed-Continuous Wavelet Transform. Geophysics, 81(4), V341-V355, doi:10.1190/GEO2015-0598.1.

Mousavi, S. M., S. P. Horton, C. A. Langston, B. Samei (2016b). Seismic Features and Automatic Discrimination of Deep and Shallow Induced-Microearthquakes Using Neural Network and Logistic Regression, Geophysical Journal International, 207(1), 29-46, doi:10.1093/gji/ggw258.

Mousavi, S. M., and C. A. Langston (2016a). Adaptive noise estimation and suppression for improving microseismic event detection, Journal of Applied Geophysics, 132, 116-124, doi:http://dx.doi.org/10.1016/j.jappgeo.2016.06.008.

Mousavi, S. M., and C. A. Langston, (2016b). Fast and novel microseismic detection using time-frequency analysis. SEG Technical Program Expanded Abstracts 2016: pp. 2632-2636. doi:10.1190/segam2016-13262278.1

Ogwari, P.O., Horton, S.P., Ausbrooks, S., (2016). Characteristics of induced/triggered earthquakes during the startup phase of the guy-greenbrier earthquake sequence in North-Central Arkansas. Seismol. Res. Lett. 87 (3), 620–630. http://dx.doi.org/10.1785/0220150252.

Perol, T., Gharbi, M., and Denolle, M. (2018). Convolutional neural network for earthquake detection and location. Science Advances, 4(2):e1700578.

Ross, Z. E., Rollins, C., Cochran, E. S., Hauksson, E., Avouac, J.-P., and Ben-Zion, Y. (2017). Aftershocks driven by afterslip and fluid pressure sweeping through a fault-fracture mesh.Geophysical Research Letters, 44(16), 8260–8267.

Ross, Z. E., Meier, M.-A., Hauksson, E., and Heaton, T. H. (2018a). Generalized Seismic Phase Detection with Deep Learning. Bulletin of the Seismological Society of America, doi: https://doi.org/10.1785/0120180080

Ross, Z. E., Meier, M.-A., and Hauksson, E. (2018b). P-wave arrival picking and first-motion polarity determination with deep learning. Journal of Geophysical Research: Solid Earth,123, 5120–5129.

Shelly, D. R., Beroza, G. C., and Ide, S. (2007). Non-volcanic tremor and low-frequency earthquake swarms. Nature, 446(7133), 305–307.

Sutskever, L.; Vinyals, O.; Le, Q. (2014). "Sequence to Sequence Learning with Neural Networks" (PDF). Electronic Proceedings of the Neural Information Processing Systems Conference. 27: 5346.2014arXiv1409.3215S.

Thireou, T.; Reczko, M. (2007). "Bidirectional Long Short-Term Memory Networks for Predicting the Subcellular Localization of Eukaryotic Proteins". IEEE/ACM Transactions on Computational Biology and Bioinformatics. 4 (3): 441–446. doi:10.1109/tcbb.2007.1015.

Yoon, C. E., O. O'Reilly, K. J. Bergen, and G. C. Beroza (2015). Earthquake detection through computationally efficient similarity search, Science Advances, 1, e1501057,http://dx.doi.org/10.1126/sciadv.1501057.

Yoon, C. E., Y. Huang, W. L. Ellsworth, and G. C. Beroza (2017). Seismicity During the Initial Stages of the Guy-Greenbrier, Arkansas, Earthquake Sequence, Journal of Geophysical Research – Solid Earth,121

Zhao, Y. and Takano, K., (1999). An artificial neural network-based seismic detector, Bull. seism. Soc. Am., 77, 670–680.

Zhu, W., and Beroza, G. C. (2018, March 8). PhaseNet: A Deep-Neural-Network-Based Seismic Arrival Time Picking Method. arXiv [physics.geo-ph].